\documentclass{article}
\usepackage{spconf,amsmath,graphicx}

\usepackage{float}
\usepackage{subfig}

\usepackage{url}
\usepackage{xspace}
\newcommand\self{\textsc{CANet}\xspace}

\title{\self: Curved Guide Line Network with Adaptive Decoder for Lane Detection}
\name{Zhongyu Yang$^{1, 2}$, Chen Shen$^{2}$, Wei Shao$^{2}$, Tengfei Xing$^{2}$, Runbo Hu$^{2}$, Pengfei Xu$^{2}$, Hua Chai$^{2}$, Ruini Xue$^{1, \ast}$\thanks{*Corresponding author}}
\address{$^{1}$School of Computer Science and Engineering, UESTC\\$^{2}$Didi Chuxing}
\begin{document}
%
\maketitle
\begin{abstract}
  Lane detection is challenging due to the complicated on-road scenarios and line deformation from different camera perspectives. Lots of solutions were proposed, but can not deal with ``corner lanes'' well. To address this problem, this paper proposes a new top-down deep learning lane detection approach, \self. A lane instance is first responded by the heatmap on the U-shaped ``\emph{curved guide line}'' at global semantic level, thus the corresponding features of each lane are aggregated at the response point. Then \self obtains the heatmap response of the entire lane through conditional convolution, and finally decodes the point set to describe lanes via adaptive decoder. The prototype is implemented with Pytorch, and evaluated against 3 well-known datasets extensively. The experimental results show that \self reaches SOTA in different metrics. Our code will be released soon.
\end{abstract}
\begin{keywords}
Lane detect, guide line, adaptive decoder
\end{keywords}
\section{Introduction}
\label{sec:intro}
Benefiting from ADAS (Advanced Driver Assistance Systems), automated driving
will be capable of controlling all aspects of driving without human
intervention. Fast, accurate lane detection is among the top challenges for
ADAS, which is generally regarded as a computer vision task. Recently, deep learning-based lane detection approaches are emerging and perform much better than traditional ways.

\begin{figure}
  \centering
  \subfloat[Model measurements.]{\label{fig:angle_chart_f1}%
    \includegraphics[width=.5\linewidth]{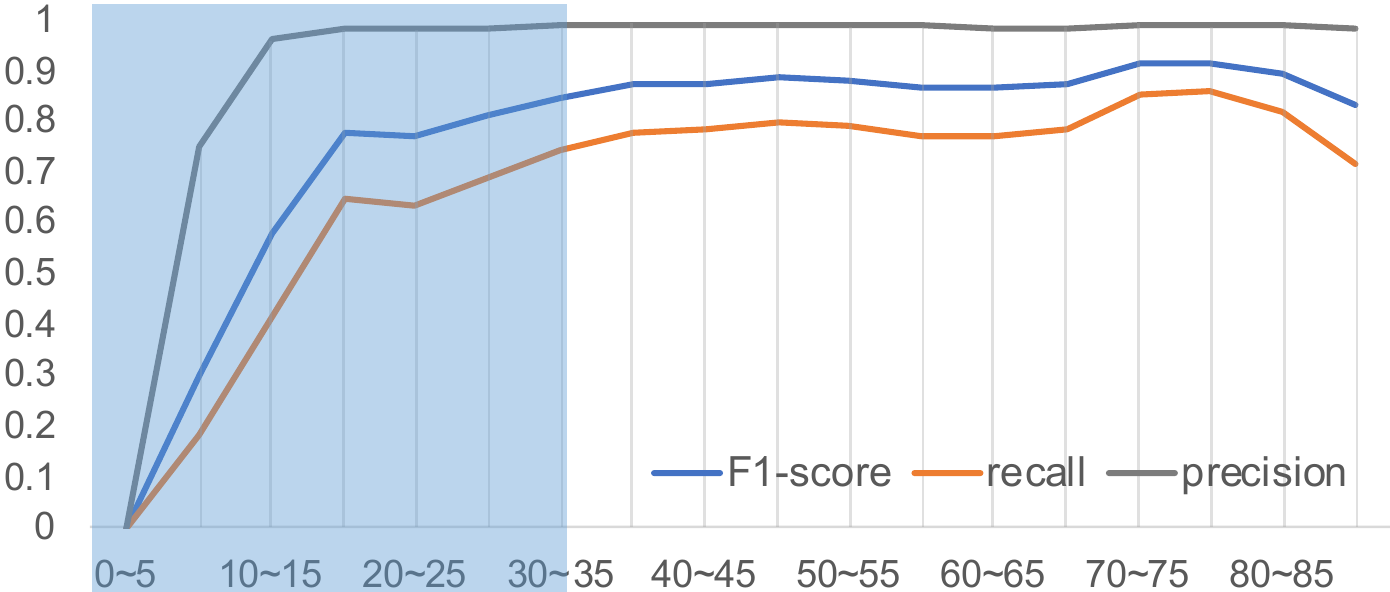}}%
  \hfill%
  \subfloat[Number of lanes.]{\label{fig:angle_chart_count}%
    \includegraphics[width=.5\linewidth]{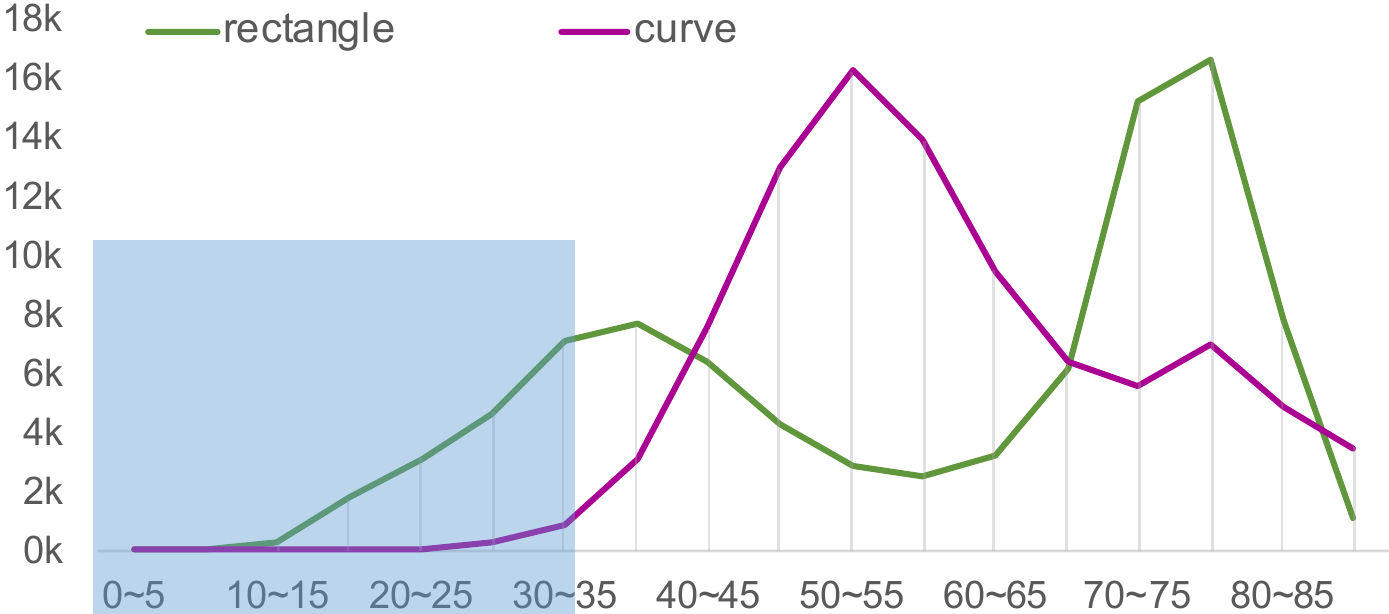}}
  \caption{Statistics for different guide lines with respect to grazing angles.}
  \label{fig:angle_chart}
\end{figure}

By leveraging the progresses in instance segmentation~\cite{yolact, blendmask, condinst}, CondLaneNet~\cite{condlanenet} devises a two-stage solution. Firstly, it uses high-level semantic features to extract the origin of each lane to represent an instance, secondly, the instance origin guides the underlying visual features to describe the shape of the instance accurately. In the first stage, the image boundary (rectangle) is the guide line for origin finding, which results in many corner lanes with small \emph{grazing angles}, making deep learning models hard to recall. Figure~\ref{fig:angle_chart_f1} illustrates the F1-score, recall, and precision for different grazing angles, which clearly shows that corner lanes of small grazing angles perform worse. In the second stage, row-wise classification is employed, but it can hardly deal with lanes that are nearly horizontal.
Besides, ordinal classification~\cite{ultra_fast} needs additional classifiers to indicate the lane range, but the range indicators and row-wise head often have inconsistent predictions at
the end of the lane, leading to anomalies such as tail flicks.

To address these problems, we propose \self, \textbf{C}urved guide line with
\textbf{A}daptive decoder \textbf{Net}work. The proposed curved guide line,
particularly an inscribed U-shaped line, can increase the grazing angles of
corner lanes for better recall. Then, Gaussian mask is used to supervise the
lane generation, enabling the network to choose either row- or column-wise
classification adaptively according to the shape of the mask as a
post-processing decoder during inference. Therefore, lane range will be
determined by the heatmap response range instead of additional classifiers,
which could avoid inconsistency otherwise.
\begin{figure*}
  \centering
  \includegraphics[width=.85\linewidth]{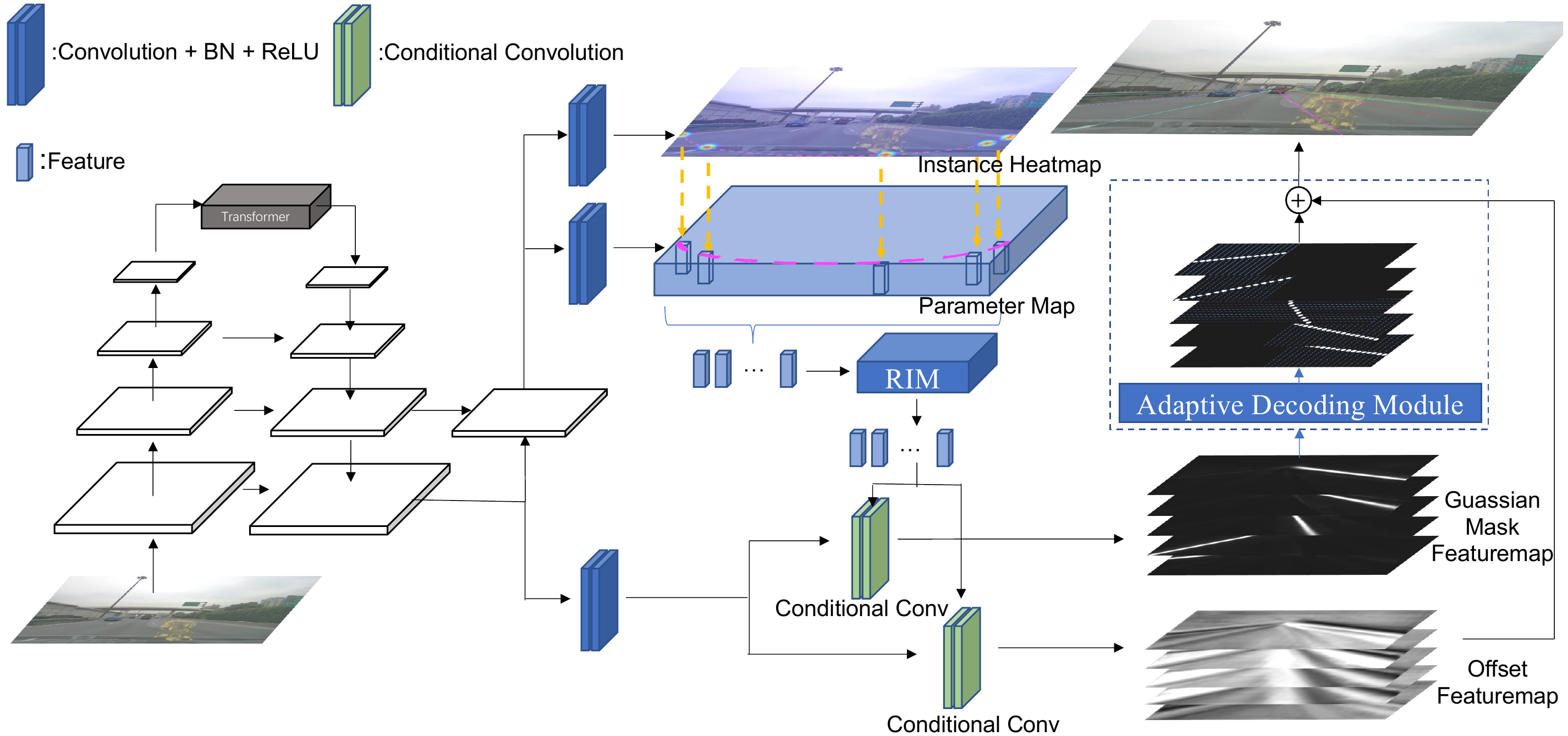}
  \caption{The structure of \self.}\label{fig:arch}
\end{figure*}

The rest of the paper is organized as follows. Section~\ref{sec:related-work}
reviews related literature and the design of \self follows in
Section~\ref{sec:methods}. Then, Section~\ref{sec:experiments} evaluates \self,
and the paper is concluded in Section~\ref{sec:concl-disc}.

\section{Related Work}
\label{sec:related-work}

Many approaches have been proposed for detecting lanes. This section discusses
the recent deep learning techniques.

\textbf{Top-down solutions}. ``Curve fitting'' approaches~\cite{liu2021end, feng2022rethinking} model
lane as a curve. However, it is hard to recognize all lanes because some
instances are not mathematically ideal. ``Anchor-based'' methods such as
LaneATT~\cite{laneatt} are similar to dense prediction in object detection, so
they can not deal with common cases like double solid lines very well.
``Row/column-wise classification''~\cite{ultra_fast, condlanenet} is widely
adopted in line-shape recognition, however they can not catch horizontal lanes.

\textbf{Bottom-up solutions}. ``Key point detection''~\cite{folo, ganet} figures
out critical points to describe the lane. The lane might be unsmooth or
incomplete in case of missing key points, which is very likely to happen for
those in the invisible parts lacking of global semantic information.
``Segmentation-based'' methods~\cite{scnn, laneaf} focus too much on pixel-level
boundaries but not capturing lane shape. Additionally, it is difficult for
segmentation to distinguish close instances because adjacent pixels usually
share similar characteristics.

\section{Methods}
\label{sec:methods}


\subsection{Network Architecture}
\label{sec:network-architecture}

As an application of instance detection, \self's architecture is inspired by many existing techniques as in Figure~\ref{fig:arch}. 
It uses ResNet~\cite{resnet} as backbone to extract image features, and uses PAFPN~\cite{pafpn} with transformer~\cite{transformer} as neck to obtain multi-scale information. In the instance extraction branch, \self
proposes using \textbf{curved guide line} to obtain the key points (i.e. origins) identifying instances, 
then acquires instance-level features through RIM (Recurrent Instance Module)~\cite{condlanenet}. Under the guidance of these features, \self calculates \textbf{Gaussian mask} and offset heatmap through the conditional convolution, and finally infers lane coordinates with \textbf{adaptive decoder}.

\subsection{Guide Line}
\label{sec:guide-line}
To the best of our knowledge, though \emph{guide lines} have been implicitly
used by some research~\cite{condlanenet} to mark the origins of lanes, we're the
first to name the boundary. If guide lines are not enforced, the origins will be
randomly distributed, which is inefficient for learning. It is straightforward
to consider the image border (a rectangle) as a guide line to restrict the
freedom of lane origins. However, in some case it is observed that the response
is very scattered, and the response peak is often indistinguishable from other
non-response points.

\begin{figure}
  \centering
  \includegraphics[width=0.7\linewidth]{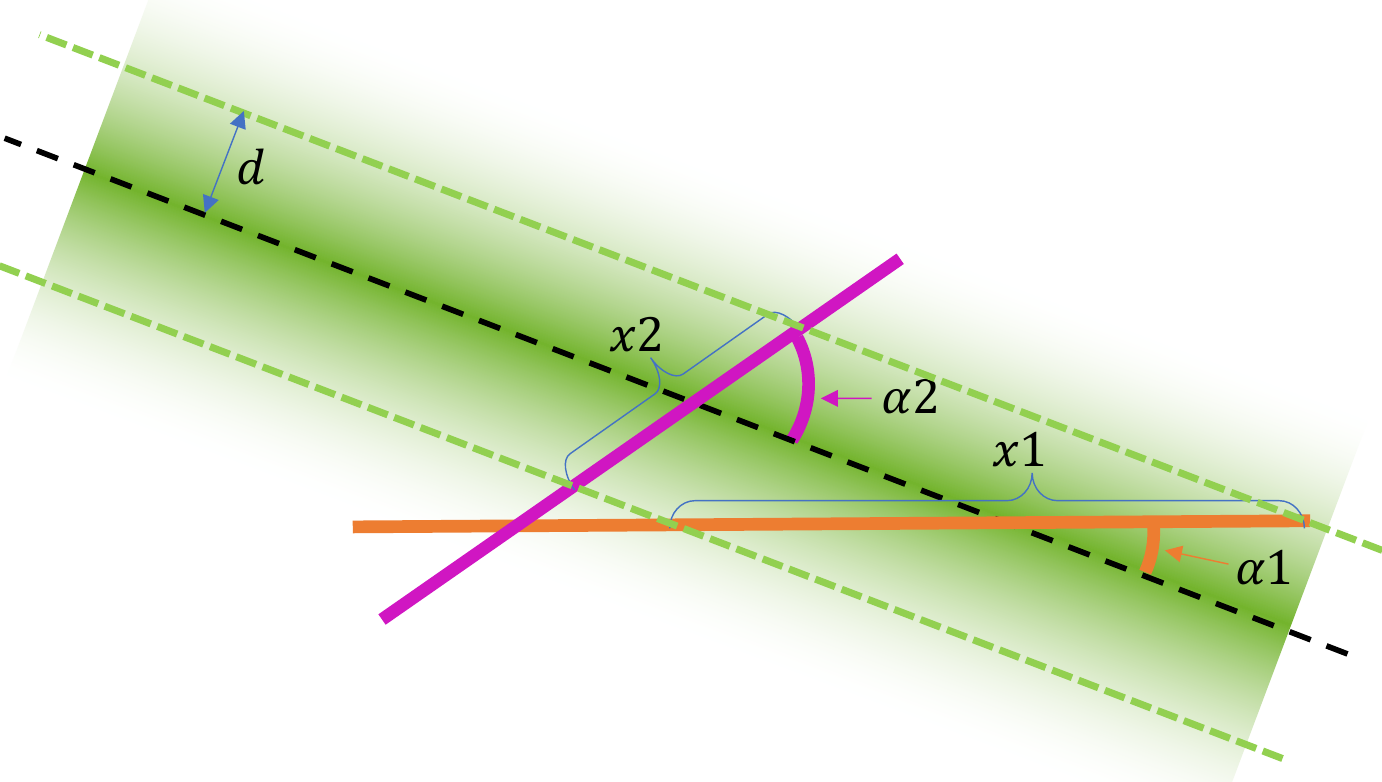}
  \caption{Response range changes with grazing angles.}
  \label{fig:analysis_grazing_angle}
\end{figure}

Figure~\ref{fig:analysis_grazing_angle} helps to understand the relation between
the response range and grazing angle. The black dotted line denotes the detected
vectorized lane (hereinafter referred to as ``vector line''), and the green
dotted lines alongside indicate the range of pixels that generate a response on
the heatmap. The greener the background is, the more positive the sample is. $d$
is the radius of the range. The tangents of two guide lines are illustrated in
orange and magenta, with grazing angles as $\alpha1$ and $\alpha2$,
respectively. Then, the relation between response range and grazing angle could
be described in Equation~\eqref{eq:response_range}. The bigger the angle (close
to 90$^{\circ}$), the tighter the response range. As the grazing angle gets
smaller, the range scatters rapidly. This reflects the results in
Figure~\ref{fig:angle_chart_f1}. Therefore, the guide line principle is to
``\emph{reduce the number of lanes of small grazing angles}''.
\begin{equation}
  \label{eq:response_range}
  x_i = \frac{2d}{\sin(\alpha_i)}; \quad i=1, 2
\end{equation}

\subsubsection{Curved Guide Line}
\label{sec:curved-guide-line}
According to the guide line principle, \self proposes to use a U-shaped curve,
the lower half is a partial ellipse and the upper half is the image side
borders. The ellipse is defined in Equation~\eqref{eq:curve}, where $w$ and $h$
are the width and height of the feature map, respectively, and $c_x$ and $c_y$
are hyper-parameters to adjust the ellipse center. Usually, $c_x$ is preferred
to 0.5, so the ellipse is in the center of the image horizontally, and the
optimal $c_y$ should be obtained by parameter experiments.
\begin{equation}
  \label{eq:curve}
    \frac{(x-c_x\cdot w)^2}{(c_x\cdot w)²} + \frac{(y-c_y \cdot h)^2}{(c_y\cdot h)^{2}} = 1
\end{equation}

Due to its incurvate property in corners, curved guide line results in bigger
grazing angles for corner lanes. Figure~\ref{fig:angle_chart_count} shows that
there are few small angles with the curved guide line. This guarantees various
model measurements will be better than that of the rectangle one.

\subsubsection{Key Points Capturing}
\label{sec:keypoints-catching}
\self follows CornerNet~\cite{cornernet} and CondLaneNet~\cite{condlanenet} to
predict a heatmap to find key points. In order to alleviate the position
inaccuracy because of downsampling, \self uses a normalized Gaussian kernel with
offset as supervision in Equation~\eqref{eq:norm}.
\begin{equation}
  \label{eq:norm}
Y_{xy} = \textrm{normalize}\left(\exp\left(-\frac{x^2+y^2}{2\sigma^2}\right)\right)
\end{equation}
where $x$ and $y$ denote the exact coordinates of each key point. Loss is
formulated as focal loss as
CondLanenet~\cite{condlanenet}, CornerNet~\cite{cornernet} and
CenterNet~\cite{centernet}.


\begin{table*}
  \caption{Comparison of different methods on CULane.}
  \label{table:culane}
  \centering
  \resizebox{.9\linewidth}{!}{%
  \begin{tabular}{lcccccccccc}
  \hline
  Method                           & Total          & Normal         & Crowded        & Dazzle         & Shadow         & No line        & Arrow          & Curve          & Cross         & Night          \\ \hline
  SCNN~\cite{scnn}                 & 71.60          & 90.60          & 69.70          & 58.50          & 66.90          & 43.40          & 84.10          & 64.40          & 1990          & 66.10          \\
  CurveLanes-L~\cite{curvelanes}   & 74.80          & 90.70          & 72.30          & 67.70          & 70.10          & 49.40          & 85.80          & 68.40          & 1746          & 68.90          \\
  LaneATT-L~\cite{laneatt}         & 77.02          & 91.74          & 76.16          & 69.47          & 76.31          & 50.46          & 86.29          & 64.05          & 1264          & 70.81          \\
  UFLDv2-M~\cite{ultra_fast}       & 75.90          & 92.50          & 74.90          & 65.70          & 75.30          & 49.00          & 88.50          & 70.20          & 1864          & 70.60          \\
  CondLaneNet-S~\cite{condlanenet} & 78.14          & 92.87          & 75.79          & 70.72          & 80.01          & 52.39          & 89.37          & 72.40          & 1364          & 73.23          \\
  CondLaneNet-M~\cite{condlanenet} & 78.74          & 93.38          & 77.14          & 71.17          & 79.93          & 51.85          & 89.89          & 73.88          & 1387          & 73.92          \\
  CondLaneNet-L~\cite{condlanenet} & 79.48          & 93.47          & 77.44          & 70.93          & \textbf{80.91} & \textbf{54.13} & 90.16          & 75.21          & 1201          & 74.80          \\ \hline
  \self-S                          & 78.46          & 93.07          & 76.59          & 70.51          & 77.82          & 52.24          & 89.39          & 72.48          & 1213          & 72.91          \\
  \self-M                          & 79.16          & 93.58          & 77.88          & \textbf{73.11} & 75.06          & 51.68          & 90.09          & 75.54          & \textbf{1176} & 73.92          \\
  \self-L                          & \textbf{79.86} & \textbf{93.60} & \textbf{78.74} & 70.07          & 79.35          & 52.88          & \textbf{90.18} & \textbf{76.69} & 1196          & \textbf{74.91} \\ \hline
  \end{tabular}}
  \end{table*}

\begin{table}
  \caption{Comparison of different methods on CurveLanes.}
\label{table:curvelanes}
\centering
\resizebox{.95\linewidth}{!}{%
\begin{tabular}{lcccr}
\hline
Method                           & F1             & Precision      & Recall         & GFlops \\ \hline
SCNN~\cite{scnn}                 & 65.02          & 76.13          & 56.74          & 328.4     \\
CurveLanes-L~\cite{curvelanes}   & 82.29          & 91.11          & 75.03          & 20.7      \\
CondLaneNet-S~\cite{condlanenet} & 85.09          & 87.75          & 82.58          & 10.3      \\
CondLaneNet-M~\cite{condlanenet} & 85.92          & 88.29          & 83.68          & 19.7      \\
CondLaneNet-L~\cite{condlanenet} & 86.10          & 88.98          & 83.41          & 44.9      \\ \hline
\self-S                          & 86.57          & 91.37          & 82.25          & 13.1      \\
\self-M                          & 87.19          & 91.53          & 83.25          & 22.6      \\
\self-L                          & \textbf{87.87} & \textbf{91.69} & \textbf{84.36} & 45.7      \\ \hline
\end{tabular}}
\end{table}

\subsection{Gaussian Mask Supervision \& Adaptive Decoder}
\label{sec:gauss-mask-superv}

\begin{figure}
  \centering
  \includegraphics[width=0.8\linewidth]{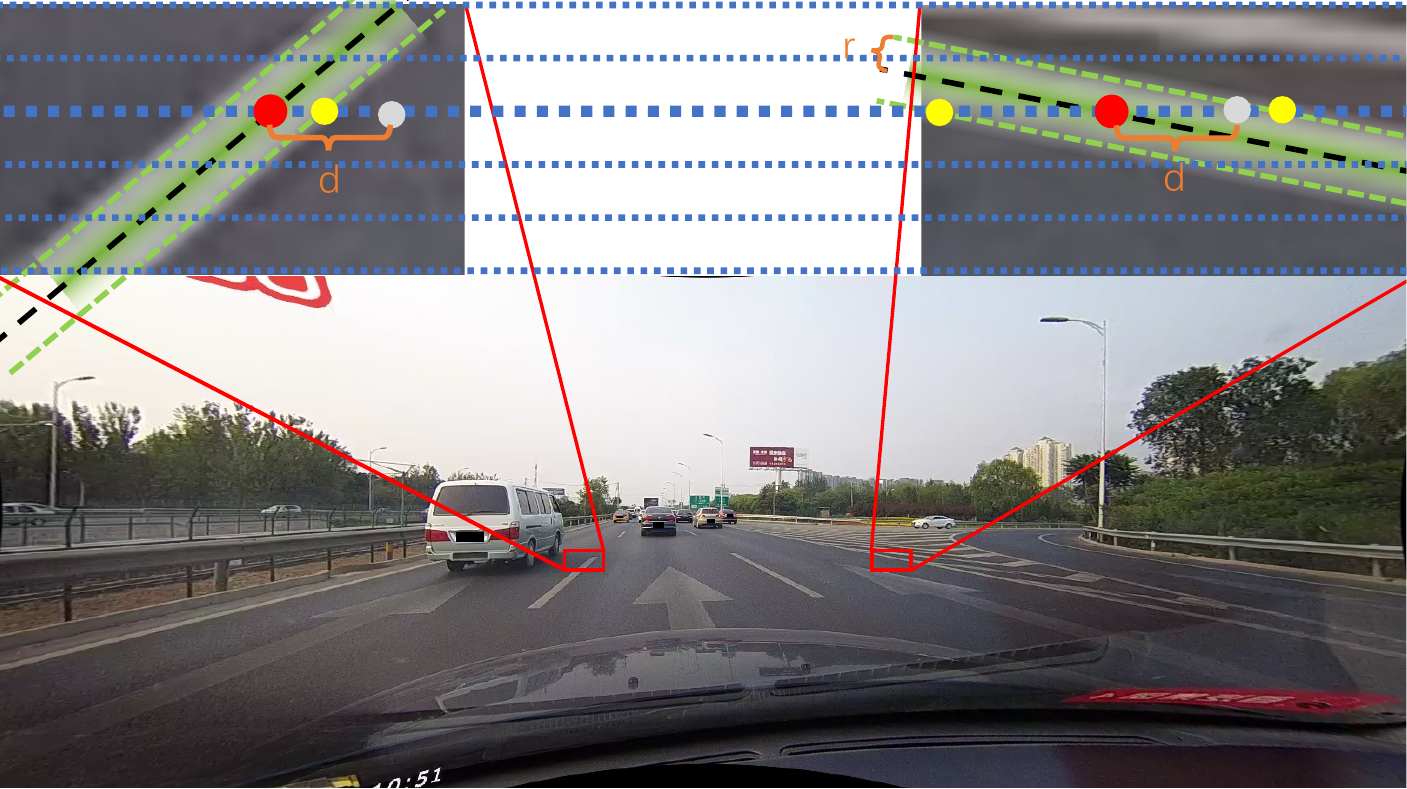}
  \caption{The distributions of positive samples for different grazing angles are
    different.}
  \label{fig:gaussian_supervision}
\end{figure}

Figure~\ref{fig:gaussian_supervision} shows a typical lane image, sharing the notations as Figure~\ref{fig:analysis_grazing_angle}. The bold blue dashed line represents a row of pixels, the red point is the intersection of the vector line and the current row, the yellow point is on the green dashed line, and the gray point is the pixel at a horizontal distance $d$ from the red one.

In traditional row-wise classification, all negative samples (yellow and gray
pixels) are weighted with the same weight. UFAST~\cite{ultra_fast} uses L1
distance to constrain expectations, making negative samples' weights
proportional to the distance from the positive sample in the same row, but it
does not consider the influence of inclination on loss. Instead, \self uses
heatmap form supervision to construct a Gaussian distribution with the vector
line as the center, to model that positive sample features decrease with
distance.

Since lanes might be horizontal or vertical, it is crucial to choose either row or column-wise classification according to the lane shape. For traditional line classification, the network cannot dynamically adjust the anchor during inference because pixels from different rows are not comparable. \self addresses this challenge with the aforementioned heatmap supervision, enabling the network to do \emph{global pixel comparison} instead, which makes it possible to change the row-wise classification in training to a dynamic post-processing operation. Then, \self is able to choose row or column-wise classification, two kinds of anchors, dynamically to the activation shape of the instance heatmap. 
Besides, the heatmap response range is the lane range, so there is no need to introduce additional classifiers that would introduce range-to-localization inconsistencies to indicate the range.

\section{Experiments}
\label{sec:experiments}

\subsection{Results}
Extensive experiments were conducted on 3 widely used lane detection datasets,
CULane~\cite{scnn}, CurveLanes~\cite{curvelanes}, and TuSimple~\cite{tusimple}.
For the sake of direct comparison, official metrics are used.

Table~\ref{table:culane} presents the results of \self compared to various models on different subsets of CULane. \self achieves the state of the art overall with 79.86 F1-score (the current best result is 79.63). Particularly, \self also delivers outstanding performance in \emph{Crowded}, \emph{Dazzle}, \emph{Cross} and \emph{Night} subsets, which are very difficult to detect. 

CurveLanes contains more difficult scenarios such as curves. Table~\ref{table:curvelanes} illustrates the results of all models on CurveLanes. \self-L achieves SOTA with 87.87 F1-score, 1.7 percentage points higher than the current best one. Actually, \self's small version \self-S surpasses the current best F1 with only 13.1/44.9$\approx$29\% computational cost. In terms of precision, \self is a bit lower than the best result (91.69 vs. 93.58), however, \self has the best recall rate, about 12.77 percentage points higher. This indicates that \self does a better job on the trade-off between recall and precision.

Compared to the other datasets, TuSimple is simpler and
Table~\ref{table:tusimple} shows that all models perform very well, and \self
provides marginal improvements on most indicators.

\subsection{Effectiveness of Curved Guide Line}

For easy comparison, we divide lanes on CurveLanes into 3 groups according to grazing angles
against rectangle guide line. Table~\ref{table:cgl_ablation} shows that
curved guide line improves recall for all groups, especially for the
smallest group. This is because curved guide line reduces the number of lanes of small grazing
angles as illustrated in Figure~\ref{fig:angle_chart_count}.

\begin{table}
    \caption{Recall for different lanes when curved guide line is disabled and enabled, respectively.}
    \label{table:cgl_ablation}
    \centering
        \begin{tabular}{cccc}\hline
                & 0°-30°        & 30°-60°        & 60°-90°          \\ \hline
        Disabled & 76.16        & 81.71         & 81.75          \\
        Enabled  & 82.51 (+6.35) & 84.58 (+2.87)  & 83.70 (+1.95)\\\hline
        \end{tabular}
\end{table}

\subsection{Ablation Study}

Table~\ref{table:ablation} presents how the two policies, curved guide line and
adaptive decoder, affect each other with CurveLanes dataset. Baseline (Line 2)
is derived from CondLaneNet-L with two updates: the downsampling factor for key
points is reduced from 16 to 8, and the rectangle guide line is enforced. These
updates make the baseline perform about 0.125 percentage points better than the
original CondLaneNet-L. In Line 3, we change the rectangle guide line to curved
guide line, and the F1-score increases by 1.036 to the baseline. Then, Gaussian
mask supervision and adaptive decoder are solely applied in Line 4 and the model
receives 1.221 increments in terms of F1. Finally, we combine both policies in
Line 5 as \self, and reach the best performance.

\begin{table}
\caption{Comparison of different methods on TuSimple.}
\label{table:tusimple}
\centering
\resizebox{0.9\linewidth}{!}{%
\begin{tabular}{lcccc}
\hline
Method                           & F1             & Accuracy       & FP            & FN            \\ \hline
SCNN~\cite{scnn}                 & 95.97          & 96.53          & 6.17          & \textbf{1.80} \\
LaneATT-L~\cite{laneatt}         & 96.06          & 96.10          & 5.64          & 2.17          \\
UFLDv2-M~\cite{ultra_fast}       & 96.22          & 95.56          & 3.18          & 4.37          \\
CondLaneNet-S~\cite{condlanenet} & 97.01          & 95.48          & 2.18          & 3.80          \\
CondLaneNet-M~\cite{condlanenet} & 96.98          & 95.37          & 2.20          & 3.82          \\
CondLaneNet-L~\cite{condlanenet} & 97.24          & 96.54          & 2.01          & 3.50          \\ \hline
\self-S                          & 97.51          & 96.56          & 2.29          & 2.68          \\
\self-M                          & 97.44          & 96.66          & 2.32          & 2.79          \\
\self-L                          & \textbf{97.77} & \textbf{96.76} & \textbf{1.92} & 2.53          \\ \hline
\end{tabular}}
\end{table}

\begin{table}
  \caption{Ablation study of the optimization policies.} 
\label{table:ablation}
\centering
\resizebox{0.9\linewidth}{!}{%
\begin{tabular}{lcccccc}
\hline
Line & Model            & F1-score       & Precision      & Recall         \\ \hline
1 & CondLaneNet         & 86.10          & 88.98          & 83.41          \\
2 & baseline            & 86.23          & 92.46          & 80.78          \\
3 & +curved guide line  & 87.26          & 91.98          & 83.01          \\
4 & +adaptive decoder   & 87.45          & \textbf{92.52} & 82.90          \\
5 & CANet               & \textbf{87.87} & 91.69          & \textbf{84.36} \\ \hline
\end{tabular}}
\end{table}

\section{Conclusion and Discussion}
\label{sec:concl-disc}
As a crucial and challenging task for automated driving, lane detection has been
widely explored from different perspectives, especially in the deep learning
era. However, the SOTA approaches are difficult to recognize corner lanes
effectively. This paper first proposes ``\emph{guide line}'' to constrain the
lane origins and suggests a U-shaped curved guide line to turn grazing angles
bigger for stable learning. By using Gaussian mask in supervision stage, the
adaptive decoder mechanism could choose between row- or column-wise
classification more intelligently, and the prediction of location and range
behave more consistently.

\bibliographystyle{ieeebib}
\bibliography{strings,refs}

\begin{thebibliography}{10}

\bibitem{yolact}
Daniel Bolya, Chong Zhou, Fanyi Xiao, and Yong~Jae Lee,
\newblock ``Yolact: Real-time instance segmentation,''
\newblock in {\em Proceedings of the IEEE/CVF international conference on
  computer vision}, 2019, pp. 9157--9166.

\bibitem{blendmask}
Hao Chen, Kunyang Sun, Zhi Tian, Chunhua Shen, Yongming Huang, and Youliang
  Yan,
\newblock ``Blendmask: Top-down meets bottom-up for instance segmentation,''
\newblock in {\em Proceedings of the IEEE/CVF conference on computer vision and
  pattern recognition}, 2020, pp. 8573--8581.

\bibitem{condinst}
Zhi Tian, Chunhua Shen, and Hao Chen,
\newblock ``Conditional convolutions for instance segmentation,''
\newblock in {\em European conference on computer vision}. Springer, 2020, pp.
  282--298.

\bibitem{condlanenet}
Lizhe Liu, Xiaohao Chen, Siyu Zhu, and Ping Tan,
\newblock ``Condlanenet: a top-to-down lane detection framework based on
  conditional convolution,''
\newblock in {\em Proceedings of the IEEE/CVF International Conference on
  Computer Vision}, 2021, pp. 3773--3782.

\bibitem{ultra_fast}
Zequn Qin, Huanyu Wang, and Xi~Li,
\newblock ``Ultra fast structure-aware deep lane detection,''
\newblock in {\em European Conference on Computer Vision}. Springer, 2020, pp.
  276--291.

\bibitem{liu2021end}
Ruijin Liu, Zejian Yuan, Tie Liu, and Zhiliang Xiong,
\newblock ``End-to-end lane shape prediction with transformers,''
\newblock in {\em Proceedings of the IEEE/CVF winter conference on applications
  of computer vision}, 2021, pp. 3694--3702.

\bibitem{feng2022rethinking}
Zhengyang Feng, Shaohua Guo, Xin Tan, Ke~Xu, Min Wang, and Lizhuang Ma,
\newblock ``Rethinking efficient lane detection via curve modeling,''
\newblock in {\em Proceedings of the IEEE/CVF Conference on Computer Vision and
  Pattern Recognition}, 2022, pp. 17062--17070.

\bibitem{laneatt}
Lucas Tabelini, Rodrigo Berriel, Thiago~M Paixao, Claudine Badue, Alberto~F
  De~Souza, and Thiago Oliveira-Santos,
\newblock ``Keep your eyes on the lane: Real-time attention-guided lane
  detection,''
\newblock in {\em Proceedings of the IEEE/CVF conference on computer vision and
  pattern recognition}, 2021, pp. 294--302.

\bibitem{folo}
Zhan Qu, Huan Jin, Yang Zhou, Zhen Yang, and Wei Zhang,
\newblock ``Focus on local: Detecting lane marker from bottom up via key
  point,''
\newblock in {\em Proceedings of the IEEE/CVF Conference on Computer Vision and
  Pattern Recognition}, 2021, pp. 14122--14130.

\bibitem{ganet}
Jinsheng Wang, Yinchao Ma, Shaofei Huang, Tianrui Hui, Fei Wang, Chen Qian, and
  Tianzhu Zhang,
\newblock ``A keypoint-based global association network for lane detection,''
\newblock in {\em Proceedings of the IEEE/CVF Conference on Computer Vision and
  Pattern Recognition}, 2022, pp. 1392--1401.

\bibitem{scnn}
Xingang Pan, Jianping Shi, Ping Luo, Xiaogang Wang, and Xiaoou Tang,
\newblock ``Spatial as deep: Spatial cnn for traffic scene understanding,''
\newblock in {\em Proceedings of the Thirty-Second AAAI Conference on
  Artificial Intelligence and Thirtieth Innovative Applications of Artificial
  Intelligence Conference and Eighth AAAI Symposium on Educational Advances in
  Artificial Intelligence}. 2018, AAAI'18/IAAI'18/EAAI'18, AAAI Press.

\bibitem{laneaf}
Hala Abualsaud, Sean Liu, David~B Lu, Kenny Situ, Akshay Rangesh, and Mohan~M
  Trivedi,
\newblock ``Laneaf: Robust multi-lane detection with affinity fields,''
\newblock {\em IEEE Robotics and Automation Letters}, vol. 6, no. 4, pp.
  7477--7484, 2021.

\bibitem{resnet}
Kaiming He, Xiangyu Zhang, Shaoqing Ren, and Jian Sun,
\newblock ``Deep residual learning for image recognition,''
\newblock in {\em Proceedings of the IEEE conference on computer vision and
  pattern recognition}, 2016, pp. 770--778.

\bibitem{pafpn}
Shu Liu, Lu~Qi, Haifang Qin, Jianping Shi, and Jiaya Jia,
\newblock ``Path aggregation network for instance segmentation,''
\newblock in {\em Proceedings of the IEEE conference on computer vision and
  pattern recognition}, 2018, pp. 8759--8768.

\bibitem{transformer}
Ashish Vaswani, Noam Shazeer, Niki Parmar, Jakob Uszkoreit, Llion Jones,
  Aidan~N Gomez, {\L}ukasz Kaiser, and Illia Polosukhin,
\newblock ``Attention is all you need,''
\newblock {\em Advances in neural information processing systems}, vol. 30,
  2017.

\bibitem{cornernet}
Hei Law and Jia Deng,
\newblock ``Cornernet: Detecting objects as paired keypoints,''
\newblock in {\em Proceedings of the European conference on computer vision
  (ECCV)}, 2018, pp. 734--750.

\bibitem{centernet}
Kaiwen Duan, Song Bai, Lingxi Xie, Honggang Qi, Qingming Huang, and Qi~Tian,
\newblock ``Centernet: Keypoint triplets for object detection,''
\newblock in {\em Proceedings of the IEEE/CVF international conference on
  computer vision}, 2019, pp. 6569--6578.

\bibitem{curvelanes}
Hang Xu, Shaoju Wang, Xinyue Cai, Wei Zhang, Xiaodan Liang, and Zhenguo Li,
\newblock ``Curvelane-nas: Unifying lane-sensitive architecture search and
  adaptive point blending,''
\newblock in {\em European Conference on Computer Vision}. Springer, 2020, pp.
  689--704.

\bibitem{tusimple}
Tusimple,
\newblock ``Tusimple lane detection benchmark, 2017,''
  \url{https://github.com/TuSimple/tusimple-benchmark},
\newblock 2017.

\end{thebibliography}

\end{document}